\pdfoutput=1

\documentclass[11pt]{article}

\usepackage[preprint]{acl}

\usepackage{times}
\usepackage{latexsym}
\usepackage{booktabs}
\usepackage[T1]{fontenc}

\usepackage[utf8]{inputenc}

\usepackage{microtype}

\usepackage{inconsolata}

\usepackage{graphicx}
\usepackage{multirow}

\usepackage{hyperref} 
\usepackage{xurl}

\title{Out of the Box, into the Clinic? Evaluating State-of-the-Art ASR for Clinical Applications for Older Adults}

\author{Bram van Dijk$^{1}$, Tiberon Kuiper$^{3}$, Sirin Aoulad si Ahmed$^{3}$, Armel Lefebvre$^{1}$, \\ {\bf Jake Johnson$^{3}$,}
{\bf Jan Duin$^{3}$,} {\bf Simon Mooijaart$^{3,4}$,} {\bf Marco Spruit$^{1,2}$} \\
         $^{1}$Department of Public Health and Primary Care, Leiden University Medical Center  \\
         $^{2}$Leiden Institute of Advanced Computer Science, Leiden University \\
         $^{3}$Center for Medicine for Older People, Leiden University Medical Center \\
         $^{4}$Department of Internal Medicine, Leiden University Medical Center \\
         Correspondence: \href{mailto:email@domain}{b.m.a.van\_dijk@lumc.nl}}

\begin{document}
\maketitle
\begin{abstract}
Voice-controlled interfaces can support older adults in clinical contexts -- with chatbots being a prime example -- but reliable Automatic Speech Recognition (ASR) for underrepresented groups remains a bottleneck. This study evaluates state-of-the-art ASR models on language use of older Dutch adults, who interacted with the \texttt{Welzijn.AI} chatbot designed for geriatric contexts. We benchmark generic multilingual ASR models, and models fine-tuned for Dutch spoken by older adults, while also considering processing speed. Our results show that generic multilingual models outperform fine-tuned models, which suggests recent ASR models can generalise well out of the box to real-world datasets. Moreover, our results indicate that truncating generic models is helpful in balancing the accuracy-speed trade-off. Nonetheless, we also find inputs which cause a high word error rate and place them in context. 
\end{abstract}

\section{Introduction}
Although there is a surge of interest in AI-driven applications like chatbots in the health domain \citep[][]{guo2024large, huo2025large}, tailoring them to groups underrepresented in AI research remains a challenge. Older populations are one example: because they often have different needs and preferences when interacting with AI \citep{van2025welzijn, klaassen2025review}, their involvement in system development is key for building clinically relevant systems in geriatrics, the field in healthcare concerned with the health of older adults. This population is increasing in size globally \citep{WHO2023}, while personnel shortages in healthcare become pressing \citep{EU2023}; yet addressing these challenges with AI warrants systems that align well with older adults.

Voice control is a key element in chatbots in geriatrics, as older individuals may struggle with small fonts, icons, and typing text in standard interfaces \citep{khamaj2025ai}. Yet, implementing voice control is not obvious, as the performance of Automatic Speech Recognition (ASR) systems depends on the representation of older adults in training data, and on their articulation, speech volume and technological literacy \citep{klaassen2025review}. Moreover, evaluation of state-of-the-art ASR systems on realistic data of older adults is lagging behind. 

In this short paper, we evaluate recent ASR models on older adults' language use in interaction with \texttt{Welzijn.AI}. This is a new digital platform for older users, that currently features a prototype chatbot to converse about clinically relevant topics like quality of life and frailty. Audio data of 10 older Dutch adults  interacting with \texttt{Welzijn.AI} were transcribed using generic multilingual ASR models (Whisper and Voxtral), and models fine-tuned specifically for Dutch or older Dutch populations (Whisper and Wav2Vec2). We find that i) generic multilingual models outperform fine-tuned models, and also that ii) truncating larger generic models helps striking a good balance between accuracy and speed. These findings also hold for a similar subset of the Mozilla Common Voice dataset \cite{ardila-etal-2020-common} we use as benchmark.

\begin{figure*}[t]
\centering
\includegraphics[width=0.80\textwidth]{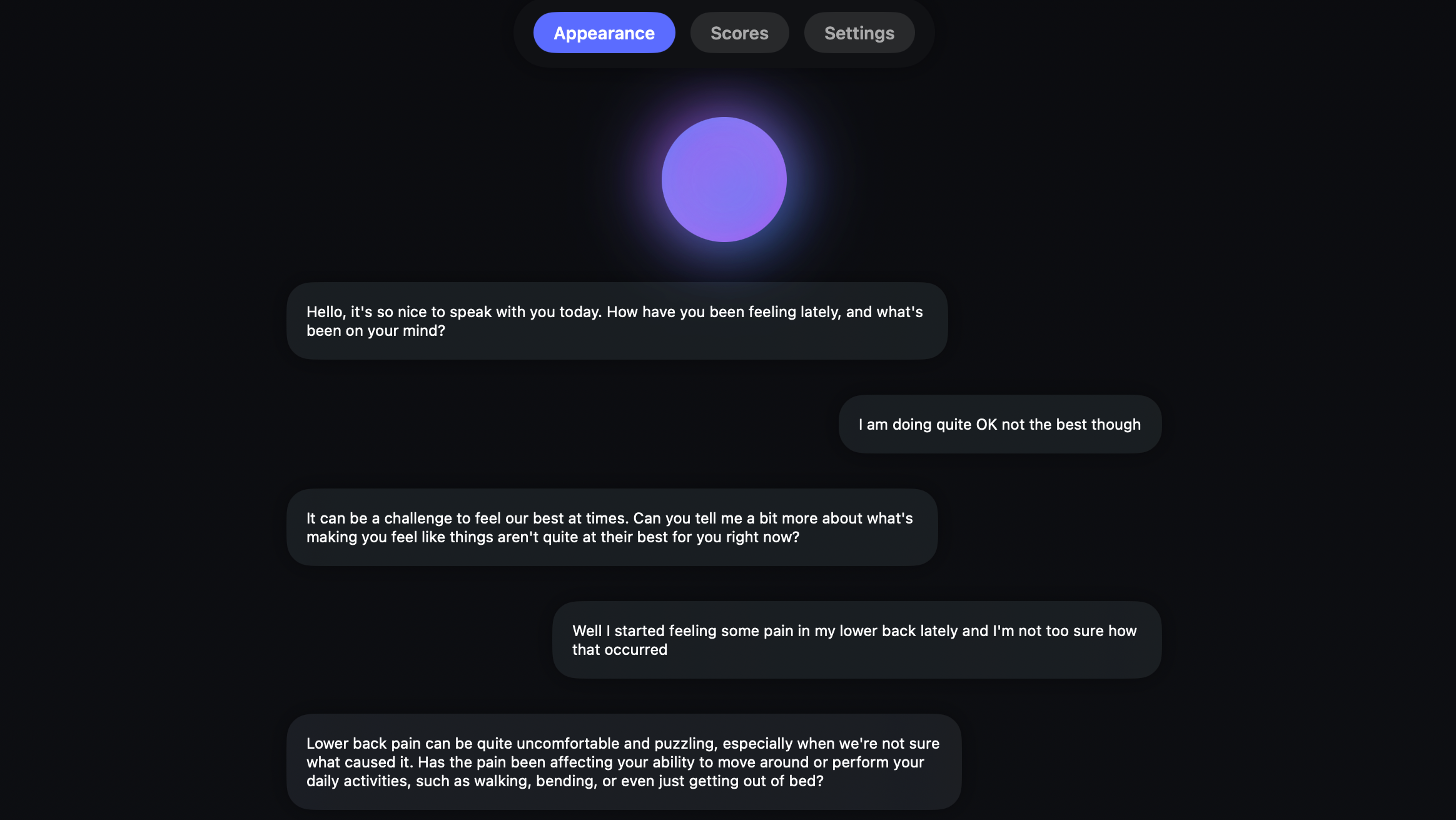}
\caption{Interface of \texttt{Welzijn.AI} with an example conversation. Users press the purple button to activate the ASR functionality and start responding, after which their speech is transcribed and rendered on the screen. Chatbot responses are read out with a text-to-speech model. The `Scores' button shows information extracted on quality of life and frailty, `Settings' allows choosing different ASR models, and `Appearance' returns the user to the conversation on display. We focus in this paper on the conversation resulting from interaction with this prototype.}
\label{fig:bea_interface}
\end{figure*}

\section{Background}
Though work on evaluating ASR models for older adults exists, extrapolating findings to realistic contexts is hard. Performance of ASR models sometimes remains implicit in downstream use, for example through evaluation of overlap of linguistic features extracted from model vs. human-generated transcripts \citep{naffah2025spoken}, or by using only part of ASR models (like the audio encoder) in predicting cognitive impairment in older persons \citep{agbavor2024multilingual}. Other work tailors ASR systems to older adults by drawing on additional databases of individual users \cite{xu2025echovoices}, which is not always feasible in a clinical context due to privacy concerns. However, \citet{xu2025echovoices} also show that fine-tuning generic multilingual models with speech from older adults increases ASR performance. The work by \citet{shekoufandeh2025improving} explores this further, by fine-tuning Whisper as recent ASR model \citep{radford2023robust} on the Dutch JASMIN-CGN dataset. This dataset includes language use of older adults in human-machine interaction settings \cite{cucchiarini2012jasmin}, which is potentially relevant to \texttt{Welzijn.AI}. 

\section{Materials and methods}
Ten older Dutch adults ($\geq$65 years) from a volunteer panel of the outpatient clinic for geriatrics at the Leiden University Medical Center (LUMC) were included. The Institutional Review Board of the LUMC approved the study and all participants provided informed consent. Participants were instructed not to share personal health information with \texttt{Welzijn.AI}, but rather to impersonate a peer. Participants interacted individually with the chatbot, with one experimenter standing by. 

An impression of \texttt{Welzijn.AI} is given in \autoref{fig:bea_interface}. Here we focus on the system's ability to support conversations; we refer for architectural details to \citet{van2025welzijn}. The chatbot was driven by the \texttt{meta-llama/Llama-3.3-70B-Instruct}\footnote{We will use Hugging Face IDs to denote AI models.} model \cite{dubey2024llama}, prompted to structure the conversation around the EQ-5D and Clinical Frailty Scale \citep{brooks1996euroqol,rockwood2005global}. These are validated geriatric instruments to assess quality of life and frailty, by retrieving information about mobility, mental wellbeing, physical independence, and so on. These instruments can be presented via surveys, in conversations with clinicians, or in our case, by a chatbot (\autoref{fig:bea_interface}).

The chatbot was used on a laptop. Interactions took 5-10 minutes and were recorded with a handheld device. The default ASR model in \texttt{Welzijn.AI} was \texttt{openai/whisper-large-v3}, which in early testing we found to work best. 

\begin{table}[h]
\centering
\small
\begin{tabular}{cc}
\hline
\textbf{Type} & \textbf{Example} \\ \hline
\textit{Orthographic}    & Hi, I am uh... feeling great today.  \\ 
\textit{Orthographic\_clean} & Hi I am feeling great today\\
\textit{Normalised} &   hi i am feeling great today \\ \hline
\end{tabular}
\caption{Examples of gold transcript types.}
\label{tab:transcriptions}
\end{table}

\begin{table}[h]
\centering
\small
\begin{tabular}{cc}
\hline
\textbf{Hugging Face ID}& \textbf{Params} \\ \hline
    \scriptsize \texttt{mistralai/Voxtral-Mini-3B-2507} & 4.68B   \\
    \scriptsize \texttt{openai/whisper-large-v2} & 1.55B   \\
    \scriptsize \texttt{openai/whisper-large-v3} & 1.55B   \\
    \scriptsize \texttt{golesheed/whisper-native-elderly-9-dutch} & 1.54B   \\
    \scriptsize \texttt{golesheed/wav2vec2-xls-r-1b-dutch-3} & 963M   \\
    \scriptsize \texttt{openai/whisper-large-v3-turbo} & 809M   \\
    \scriptsize \texttt{openai/whisper-medium} & 769M   \\
    \scriptsize \texttt{openai/whisper-small} & 244M   \\
    \hline
\end{tabular}
\caption{Models used for our ASR experiments.}
\label{tab:models}
\end{table}

\begin{table*}[t]
\centering
\small
\begin{tabular}{c|ccc|c|ccc|c}
\toprule
\textbf{Hugging Face ID}& \multicolumn{3}{c|}{\textbf{WER Welzijn.AI}} &
& \multicolumn{3}{c|}{\textbf{WER Common Voice}} \\
\cmidrule(lr){2-5} \cmidrule(lr){6-9}
& Orth. & Orth\_c. & Norm. & \textit{Time}
& Orth. & Orth\_c. & Norm. & \textit{Time} \\
\midrule
\scriptsize \texttt{mistralai/Voxtral-Mini-3B-2507} & .17 & .11 & .09 & 3.75 & .05 & .04 & .04 & 3.72 \\
\scriptsize \texttt{openai/whisper-large-v2} & .19 & .12 & .10 & 4.69 & .05 & .04 & .04 & 4.24 \\
\scriptsize \texttt{openai/whisper-large-v3} & \textbf{.12} & \textbf{.07} & \textbf{.06} & 3.41 & \textbf{.04} & \textbf{.03} & \textbf{.03} & 3.71 \\
\scriptsize \texttt{golesheed/whisper-native-elderly-9-dutch} & .40 & .22 & .14 & 4.59 & .29 & .18 & .07 & 4.07 \\
\scriptsize \texttt{golesheed/wav2vec2-xls-r-1b-dutch-3} & .49 & .37 & .36 & \textbf{.99} & .30 & .19 & .19 & \textbf{.89} \\
\scriptsize \texttt{openai/whisper-large-v3-turbo} & .16 & .10 & .08 & 1.43 & .06 & .04 & .04 & 1.47 \\
\scriptsize \texttt{openai/whisper-medium} & .19 & .13 & .11 & 2.35 & .07 & .06 & .06 & 2.48 \\
\scriptsize \texttt{openai/whisper-small} & .26 & .18 & .17 & 1.11 & .12 & .10 & .10 & 1.19 \\
\bottomrule
\end{tabular}
\caption{Word Error Rate (WER) is the edit distance between prediction and reference (sum of substitutions, deletions, and insertions), divided by the length of the reference, so here denotes the average number of errors per reference word, for Orthographic (\textit{Orth.}), Orthographic\_clean (\textit{Orth\_c.}), and Normalised (\textit{Norm.}) gold transcripts. Processing time (\textit{Time}) in average seconds per input. Best results in bold.}
\label{tab:results}
\end{table*}

Recorded user speech was after the interactions separated from chatbot responses and segmented using \texttt{PyDub}\footnote{\url{https://github.com/jiaaro/pydub}} and \texttt{pyannote} \citep{Bredin23}. We obtained 199 segments with an average length of 3.4 seconds, due to the turn-taking nature of the conversation. Our sample totalled 11 min. and 15 sec., so is small, but still valuable as data from clinical contexts is challenging to obtain. Besides chatbot data, we also drew 200 random samples from the Common Voice dataset \citep{ardila-etal-2020-common} of Dutch older individuals ($\geq$ 60 years), which totals 17 min., and 37 sec. As this data concerns written text read out loud, it should intuitively be an easier benchmark for ASR models.

For obtaining gold standard (i.e. human) reference transcriptions, segments were transcribed with the default \texttt{openai/whisper-large-v3} model, and subsequently corrected by the first author. Since the Word Error Rate (WER) as standard metric in the ASR field is sensitive to fillers, capitalisation, and punctuation, and since choosing the `right' reference depends on the use case, we created three types of human and model transcriptions (via rule-based postprocessing) as visible in \autoref{tab:transcriptions}: \textit{orthographic} (including fillers, capitalisation, and punctuation) \textit{orthographic\_clean} (only capitalisation, as also used in \autoref{fig:bea_interface}), and \textit{normalised} (no fillers, capitalisation or punctuation). Orthographic transcription is useful in that it provides additional structure, though speech content is often arguably sufficiently preserved in normalised transcriptions, with orthographic\_clean transcription striking a balance between the strictest and most flexible evaluation scenarios. 

\autoref{tab:models} shows the ASR models included in our experiments. As can be seen, we focus on models from the \texttt{Whisper} model family as the current standard in the field, and include \texttt{mistralai/Voxtral-Mini-3B-2507} \citep{liu2025voxtral} as new potential competitor. We included \texttt{golesheed/whisper-native-elderly-9-dutch} as the Whisper model fine-tuned for older Dutch adults by \citet{shekoufandeh2025improving}, and we also included \texttt{golesheed/wav2vec2-xls-r-1b-dutch-3}, a work-in-progress Wav2Vec2 model \citep{baevski2020wav2vec}  fine-tuned for general Dutch, as older but potentially fast contender. We note that these fine-tuned models do not output fillers, capitals or punctuation by default, hence evaluate them with normalised transcripts. 

Our motivation for this set of ASR models is that they are all small (compared to Voxtral's 24B variant for example) and open weights, hence suitable for local/private downstream applications. We take processing time into account since a good balance between accuracy and speed is key in many applications, so include models of different sizes.

Our code for the ASR pipeline is available.\footnote{\url{https://github.com/bma-vandijk/asr_pipelines}} All experiments were carried out on a Macbook Pro M1 16GB using the \texttt{Hugging Face} ecosystem for ASR and \texttt{PyTorch}'s MPS GPU acceleration backend. Due to privacy restrictions we cannot share recordings nor transcripts of the interactions.

\begin{table*}[h]
\centering
\scriptsize
\begin{tabular}{ccll|c}
\hline
\textbf{\#} & \textbf{Model} & \textbf{Prediction} & \textbf{Reference} & \textbf{WER} \\ \hline

\multirow{2}{*}{1} & \multirow{2}{*}{\texttt{openai/whisper-large-v2}} & hartelijk bedankt voor het kijken en tot de volgende keer & gaat wel & \multirow{2}{*}{5} \\
 &  & \textit{thank you cordially for watching and until next time} & \textit{it’s okay} & \\[0.3em]

\multirow{2}{*}{2} & \multirow{2}{*}{\texttt{golesheed/whisper-native-elderly-9-dutch}} & poet hem in oranje & goedemiddag & \multirow{2}{*}{4} \\
 &  & \textit{loot him in orange} & \textit{good afternoon} & \\[0.3em] 

\multirow{2}{*}{3} & \multirow{2}{*}{\texttt{openai/whisper-large-v2}} & ik denk dat het weer zo is als het altijd is & ongeveer zoals altijd & \multirow{2}{*}{3.33} \\
 &  & \textit{I think that it is as it always is} & \textit{roughly as usual} & \\[0.3em] 

\multirow{2}{*}{4} & \multirow{2}{*}{\texttt{openai/whisper-medium}} & ik ben benieuwd & goedemiddag & \multirow{2}{*}{3} \\
 &  & \textit{I am curious} & \textit{good afternoon} & \\[0.3em] 

\multirow{2}{*}{5} & \multirow{2}{*}{\texttt{openai/whisper-small}}& voor de wereld & goedemiddag & \multirow{2}{*}{3} \\
 &  & \textit{for the world} & \textit{good afternoon} & \\[0.3em] 

\multirow{2}{*}{6} & \multirow{2}{*}{\texttt{mistralai/Voxtral-Mini-3B-2507}} & groen de medaillon & goedemiddag & \multirow{2}{*}{3} \\
 &  & \textit{green the medallion} & \textit{good afternoon} & \\[0.3em] 

\multirow{2}{*}{7} & \multirow{2}{*}{\texttt{golesheed/wav2vec2-xls-r-1b-dutch-3}} & goede midda & goedemiddag & \multirow{2}{*}{2} \\
 &  & \textit{good midda} & \textit{good midday} & \\[0.3em] 

 \multirow{2}{*}{8} & \multirow{2}{*}{\texttt{openai/whisper-large-v3-turbo}} & ja het is heel erg goed & weer te ingewikkeld & \multirow{2}{*}{2} \\
 &  & \textit{yes it is very well} & \textit{again too complicated} & \\[0.3em] 

 \multirow{2}{*}{9} & \multirow{2}{*}{\texttt{openai/whisper-medium}} & bedankt voor het kijken & gaat wel & \multirow{2}{*}{2} \\
 &  & \textit{thank you for watching} & \textit{it's okay}& \\[0.3em]

 \multirow{2}{*}{10} & \multirow{2}{*}{\texttt{mistralai/Voxtral-Mini-3B-2507}} & het zelf gaat het & hetzelfde uiteraard & \multirow{2}{*}{2} \\
 &  & \textit{it self goes it} & \textit{the same of course} & \\[0.3em]
 
\end{tabular}
\caption{Example inputs with high WER at the sample level. English (literal) translations in italics.}
\label{tab:errors}
\end{table*}

\section{Results}
\begin{figure}[t]
\centering
\includegraphics[width=0.5\textwidth]{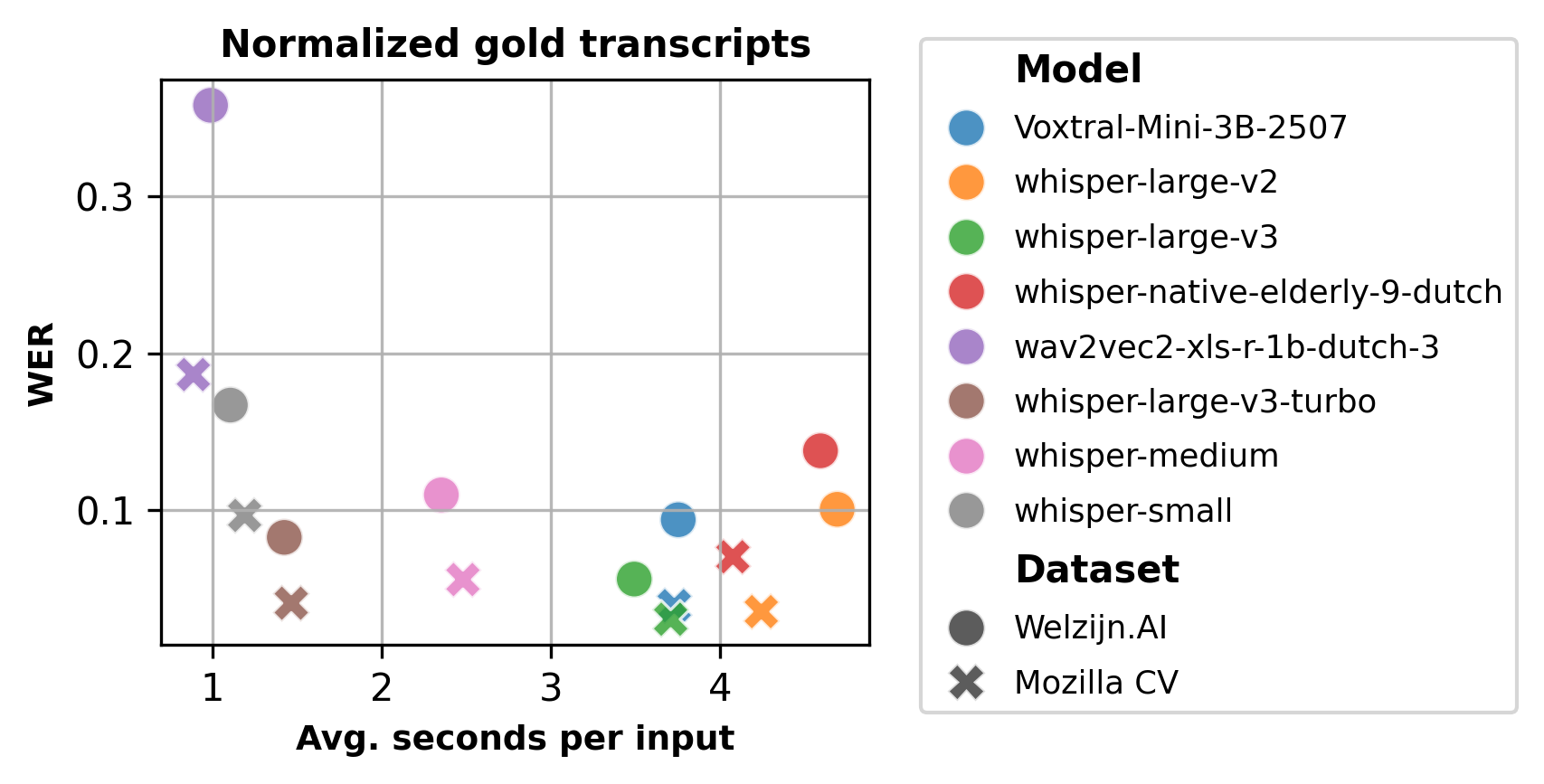}
\caption{Overview of accuracy vs. processing time.}
\label{fig:wer_time_plot}
\end{figure}

WER and processing time per model are given in \autoref{tab:results}. Regarding WER, on both the \texttt{Welzijn.AI} and Common Voice datasets, \texttt{openai/whisper-large-v3} as generic multilingual model outperforms all other models, also regarding normalised transcripts as most flexible evaluation scenario. In terms of processing time, on both datasets, the Wav2Vec2 model fine-tuned on Dutch language use (not specifically older language users) (\texttt{golesheed/wav2vec2-xls-r-1b-dutch-3}) is the fastest, but also the least accurate.\\ 
\indent For chatbot systems like \texttt{Welzijn.AI}, understanding the trade-off between WER and processing time is crucial, given that in chatbots other components also impose processing time, and seconds may greatly impact the perceived quality of the experience. We visualise performance w.r.t. normalised transcripts in \autoref{fig:wer_time_plot}. Here we see that \texttt{openai/whisper-large-v3-turbo}, which is essentially \texttt{openai/whisper-large-v3} with a truncated decoder 1/8 its size, strikes the best balance. The `nearest' improvement in WER concerns \texttt{openai/whisper-large-v3}, which is about three times slower, while the `nearest' improvement in processing time comes from \texttt{openai/whisper-small}, at the expense of more than twice its WER. \autoref{fig:wer_time_plot} also shows that for all models, the Common Voice dataset is easier regarding WER, though not all models process these data faster. For Common Voice data the same observations hold regarding models that are the `nearest' improvements in WER and processing times compared to \texttt{openai/whisper-large-v3-turbo}: they lead to considerable drops in speed or accuracy respectively. Our findings align with earlier work that shows that overall, larger ASR models perform better on commonly used datasets compared to smaller models \cite{atwany2025lost}, though they also note that larger numbers of parameters eventually yield diminishing returns. This finding helps understand why \texttt{openai/whisper-large-v3-turbo} shows a relatively small performance drop while still being about 50\% smaller than \texttt{openai/whisper-large-v3}.         

\subsection{Error analysis}
To disclose common pitfalls in using ASR models on Dutch, we provide ten examples of the 50 predictions with highest WER ($\geq$ 2) in \autoref{tab:errors}, which as expected come only from \texttt{Welzijn.AI} data. Errors in predictions in deep neural ASR models like Whisper can be categorised in various ways. \textit{Mishearings} could be induced by ambiguous or unclear phonemes in the input, or confusion of phonemes by the model (e.g. `this guy' vs. `the sky'). \textit{Hallucinations} are errors where the prediction has no semantic or phonetic relation to the reference. \textit{Looping} is when a model keeps repeating previously recognized speech (`Welcome to Amsterdam to Amsterdam to Amsterdam'). Furthermore, it is also known that deep neural ASR models are sensitive to non-speech audio signals in the background caused by e.g. objects or animals \cite{baranski2025investigation}, which is not obvious to trace in the predictions, but in live interaction settings like \texttt{Welzijn.AI} something to take into account.                 

Examples 2, 3, 6, 7 and 10 seem cases of mishearings, where 2 and 6 have at least some phonetic (but not semantic) alignment with the reference; examples 3, 7 and 10 also have some semantic overlap. Examples 1, 4, 5, 8, 9 have no clear phonetic or semantic link with the reference thus qualify as hallucinations; 1 and 9 are probably frequency effects from Whisper's training data (video transcriptions) \citep{baranski2025investigation}, meaning the models prioritize patterns in the training distribution over the actual audio input.  All in all, our examples suggest high WER is not limited to just a few types of models, which aligns with earlier documented unpredictability in state-of-the-art ASR models across the board \citep{koenecke2024careless, atwany2025lost} , and this finding should inform their development and deployment in high-stakes contexts.

\section{Discussion and conclusion}
We evaluated state-of-the-art ASR models on transcribing language use of older adults interacting with the \texttt{Welzijn.AI} chatbot, which was designed for geriatrics. We included various generic multilingual models as well as models fine-tuned on language use of older Dutch adults and general Dutch. We found that on both \texttt{Welzijn.AI} and Common Voice data, generic multilingual models perform better than fine-tuned models, with \texttt{openai/whisper-large-v3} as best model achieving WERs of .06 and .12 for normalised and orthographic transcriptions of realistic \texttt{Welzijn.AI} conversations. Interestingly, its truncated variant \texttt{openai/whisper-large-v3-turbo} struck the best balance between accuracy and processing speed, the latter being crucial in chatbot systems used in real-time. This is useful from a systems development perspective, since truncated models may perform well out of the box, without the need for training smaller architectures from scratch, or for additional data for group- or task-specific fine-tuning. Future work should further support this claim by using larger samples across settings. 

To put our results in perspective, other evaluations of state-of-the-art ASR models on the entire Dutch subset of the Common Voice dataset reported WERs of .06 by \texttt{mistralai/Voxtral-Mini-3B-2507} \citep{liu2025voxtral}, and .04 by \texttt{openai/whisper-v3-large}, assuming the strictest evaluation scenario (orthographic).\footnote{Results of latest model reported on \url{https://github.com/openai/whisper}, August 6 2025.} We obtained similar results for our subset of Common Voice spoken by Dutch Adults: .04 for the same Whisper and .05 for the same Voxtral model. Hence, it seems that for generic multilingual models, changing the target population does not imply large performance degradation if the task is straightforward (reading text out loud).    

Still, our \texttt{Welzijn.AI} data is conversational, hence results are harder to put in perspective. Though the WER on orthographic transcripts for \texttt{Welzijn.AI} data triples for the same Voxtral and Whisper models (.17 and .12 respectively) , given the different nature of read speech and conversational language, this is not a dramatic loss of performance. Recent work has reported WERs for a variety of datasets transcribed with our best performing model \texttt{openai/whisper-v3-large}, as large as .32 for English speech recordings in home environments (BERSt), and .23 for English meeting recordings (AMI) \cite{atwany2025lost}.

We also attempted to categorize WER error types. We saw that mishearings were as frequent as hallucinations. Hallucinations, however, are potentially more problematic for systems like \texttt{Welzijn.AI}, as for a user who is unable to make sense of the resulting transcription, trust will erode faster compared to a mishearing, which still has some semantic or phonetic similarity. 

Strategies to improve WER and mitigate hallucinations, could include more independent language modelling components that take specific contexts into account. When the audio input for the decoder is noisy, its language prior generates a transcript based on its own distribution instead of the input, which may well be out-of-context. Mitigation could involve generating candidate predictions and evaluating their likelihoods from the perspective of a domain-specific language model, or combining the decoder's predicted token probabilities directly with the prediction of a context-specific language model \citep[see also][]{zhou2025improving}. Hence, some exciting work remains for making real-world impact with recent ASR models. 

\section{Limitations}
Though dedicated GPUs in high performing computing clusters have higher bandwidths and are faster, hence the default choice in experiments like ours, unified architectures (such as provided by Apple's Silicon M series) are receiving more attention nowadays due to their benefits in terms of latency, energy consumption, and system footprint, and are recognized as efficient and competitive alternatives to dedicated GPUs \citep{hubner2025apple,kenyon2022apple}. So faster processing times could probably be attained by using dedicated GPUs or by using a more recent M-chip. Still, we anticipate a scenario where one device hosts multiple smaller AI models to do different tasks, for which our setup can provide a good lower bound.

Also, though we tried to make comparison fair for different ASR models by evaluating with different kinds of transcripts, developing further transcript normalisations to take `acceptable errors' into account, e.g. writing numbers in digits or letters (`8', `eight'), were beyond the scope of the current work. This means that there can be some noise in our performance estimates.

\bibliography{custom}

\appendix

\end{document}